\documentclass[preprint,12pt]{elsarticle}

\usepackage{graphicx}
\usepackage{amssymb}
\usepackage{textcomp}
\usepackage[colorlinks]{hyperref}
\hypersetup{colorlinks,citecolor=blue}

\biboptions{round,authoryear}

\renewcommand\footnotemark{}

\makeatletter
\def\ps@pprintTitle{%
 \let\@oddhead\@empty
 \let\@evenhead\@empty
 \def\@oddfoot{}%
 \let\@evenfoot\@oddfoot}
 \def\footnote{\gdef\@thefnmark{}\@footnotetext}
\makeatother

\usepackage[hang,flushmargin]{footmisc}

\begin{document}

\begin{frontmatter}


\title{\textbf{Gap Analysis of Natural Language Processing Systems with respect to Linguistic Modality}}



\author{Vishal Shukla \footnote{V. Shukla\\ Systems Engineer \\
              \textit{Infosys}, Bangalore, India. \\
              Contact no.: +91-97394 99580\\
              E-mail: vishal\_shukla04@infosys.com}}



\begin{abstract}
Modality is one of the important components of grammar in linguistics. It lets speaker to express attitude towards, or give assessment or potentiality of state of affairs. It implies different senses and thus has different perceptions as per the context. This paper presents an account showing the gap in the functionality of the current state of art Natural Language Processing (NLP) systems. The contextual nature of linguistic modality is studied. In this paper, the works and logical approaches employed by Natural Language Processing systems dealing with modality are reviewed. It sees human cognition and intelligence as multi-layered approach that can be implemented by intelligent systems for learning. Lastly, current flow of research going on within this field is talked providing futurology.
\end{abstract}

\begin{keyword}
modality \sep NLP \sep gap \sep shortcomings \sep multi-layered \sep learning


\end{keyword}

\end{frontmatter}

\section{Introduction}
\label{intro}
Research in NLP is gaining interest as its applications are becoming more significant. Natural language is highly ambiguous and understanding it effectively can be considered as a primary concern. Modality in language is associated with contextual understanding and implied perceptions. Defining modality from a computational linguistics perspective is somewhat difficult because several concepts are used to refer to phenomena that are related to modality, depending on the task at hand and the specific singularities that the speaker addresses. There are different senses that can be articulated by modality. Understanding precise sense conveyed is important.

The major tasks involved in dealing with modality in text are: detecting the occurrence of modality, categorizing the type of modality and perceiving the sense conveyed through it. Parsers and language processing tools identifies the occurrence of modals by Part of Speech (POS) tagging. Talking about type of modality, in the literature, standard classification of types is not available as various types are defined. But we consider the classification broadly as in two types: epistemic and deontic. According to \citet{palmer2014modality}, epistemic is used by the speakers “to express their judgment about the factual status of the proposition”. Whereas deontic modality is concerned with the speaker’s directive attitude towards an action to be carried out. It relates to obligation or permission and to conditional factors “that are external to the relevant individual”. The senses of necessity and possibility are incorporated by epistemic type while those of permission and obligation by deontic. Modality type categorization and the sense recognition has been carried out by annotation approaches.

This paper will closely observe the methods applied for the abovementioned tasks and also review the works in the subject field. It will then highlight shortcomings of the state of art natural language systems in context to linguistic modality that forming the ‘gap’ in their functionality. Latter sections will be covering the directions for the scope of improvement in the same context.

\section{Context Review}
\label{review}
\indent As tagging and annotating are the two key aspects, let’s summarize them regarding to the language processing. The process of tagging involves assigning tags to each word in the corpus corresponding to the part of speech that it embodies. The Part of Speech tagging is an essential subtask in language processing and is very useful for subsequent phases like syntactic parsing. The tags into which a token can possibly classified depends on the tagset adopted for this task. That is, it depends upon the Treebank taken into use which defines the directory of the tags. 

Annotation, on the other hand, uses machine learning approach. It uses pre-annotated corpus to learn and annotate plain text. The annotations made are markups usually representing extended features. The extended features may include polarity, certainty, subjectivity, sense and type of modality, event mentions, etc. POS tagging is also carried out with machine learning methods in a similar way as annotation \citep{manning1999foundations,ratnaparkhi1996maximum,toutanova2000enriching}.

\subsection{Detecting occurrence}
\indent A number of different expressions in language can have modal meanings. \citet{von2006modality} discusses on a subset of variety of modal expressions. Taking into account the tagging output of Stanford NLP parser of few sample sentences with different expressions, some points will be highlighted.

In case of modal auxiliaries used, the parser tags the same with ‘MD’ i.e. modal. \\
\begin{enumerate}
\item{Sandy must be home. \\
Sandy/NNP\hspace{15pt} must/MD\hspace{15pt} be/VB\hspace{15pt} home/NN\hspace{15pt} ./. \\}
\end{enumerate}

Thus any explicit occurrence of modal auxiliaries such as \textit{must, should/shall, might, may, could/can} will be detected well and clear.

Whereas semi-modals like \textit{has to, need to, ought to} follow different treatment. Sometimes \textit{ought to} is considered as modal and sometimes as semi-modal because of difference in its syntax. Anyhow it is tagged as a modal. Rest of the semi-modals are not tagged as modal; but auxiliary verbs and structure of sentence can be identified by parsers.\\

\begin{enumerate}
\setcounter{enumi}{1}
\item{Sandy has to be home.\\
Sandy/NNP\hspace{15pt} has/VBZ\hspace{15pt} to/TO\hspace{15pt} be/VB\hspace{15pt} home/NN\hspace{15pt} ./.\\}

\item{Sandy ought to be home.\\
Sandy/NNP\hspace{15pt} ought/MD\hspace{15pt} to/TO\hspace{15pt} be/VB\hspace{15pt} home/NN\hspace{15pt} ./.\\}
\end{enumerate}

Apart from modal auxiliaries and semi-modals, modal meaning can also be conveyed using adverbs (perhaps, probably, etc.), nouns (possibility, necessity, etc.), adjectives (bound, certain, etc.), and also conditional constructs (if.. then..). \\

\begin{enumerate}
\setcounter{enumi}{3}
\item{It is far from necessary that Sandy is home. \\
It/PRP\hspace{15pt} is/VBZ\hspace{15pt} far/RB\hspace{15pt} from/IN\hspace{15pt} necessary/JJ\hspace{15pt} that/IN\hspace{15pt}\\ Sandy/NNP\hspace{15pt} is/VBZ\hspace{15pt} home/NN\hspace{15pt} ./.\\}

\item{There is a slight possibility that Sandy is home. \\
There/EX\hspace{15pt} is/VBZ\hspace{15pt} a/DT\hspace{15pt} slight/JJ\hspace{15pt} possibility/NN\hspace{15pt} that/IN\hspace{15pt} Sandy/NNP\hspace{15pt} is/VBZ\hspace{15pt} home/NN\hspace{15pt} ./.\\}

\item{Perhaps Sandy is home. \\
Perhaps/RB\hspace{15pt} Sandy/NNP\hspace{15pt} is/VBZ\hspace{15pt} home/NN\hspace{15pt} ./. \\}

\item{If the light is on, Sandy is home. \\
If/IN\hspace{15pt} the/DT\hspace{15pt} light/NN\hspace{15pt} is/VBZ\hspace{15pt} on/IN\hspace{15pt} ,/,\hspace{15pt} Sandy/NNP\hspace{15pt} is/VBZ\hspace{15pt} home/NN\hspace{15pt} ./.\\}
\end{enumerate}

Although, the advances in calibration of parsers has improved the ability to tag words accurately, but above certain point, the mechanism seem to become insufficient to gather underlying information that is not superficial or apparent. 

\subsection{Type Categorization and Sense Perception}

\indent Categorization of type of modality in text and the identification of sense conveyed can be done by developing annotation schemes. There are works that accommodate annotating of those features, but not necessarily they are organized with study of language’s perspective; which makes it difficult to summarize and separate out the relevant points of interest. Upon reviewing the literature, it can be seen that various annotating schemes has been constructed over time for marking annotations of different components.

\citet{baker2012modality} described a modality/negation (MN) annotation scheme which isolates three components of modality and negation: a \textit{trigger} (that is source of modality or negation), a \textit{target} (action associated with modality or negation) and a \textit{holder} (the experiencer of modality). Moreover they have constructed MN lexicon and two automated MN taggers using the annotation scheme.

\citet{ruppenhofer2012yes} presents annotation scheme that annotates type of English modals in MPQA corpus. The modal verbs targeted were \textit{can/could, may/might, must, ought, shall/should}. The annotation involved categorization of the modals in to six types: \textit{epistemic, deontic, dynamic, optative, concessive, conditional}.

\citet{pakray2012automatic} experimented on QA4MRE data sets to identify modality and negation in text and assign labels \textit{mod, neg, neg-mod, none} for occurrences of modal, negation, modality and negation, and absence of both modality and negation respectively. And the detected modals were categorized into \textit{epistemic} and \textit{deontic}. 

\citet{hendrickx2012modality} presents a scheme for annotation of modality in Portuguese, using MMAX2 tool. The components annotated were: \textit{trigger} (the element conveying the modal value), \textit{target} (the expression in the scope of the trigger), \textit{source of the event mention} (speaker or writer), and \textit{source of the modality} (agent or experiencer). Also, for trigger, two attributes were specified: \textit{modal value} and \textit{polarity}. They stated thirteen different types into which the modal value could be categorized.

\citet{rubinstein2013toward} proposed fine-grained annotation approach utilizing MPQA corpus and MMAX2 tool. It was said to be fine-grained as it extends some of the previous works with a number of novel features to improve detection and interpretation of modals in text. The features annotated were: \textit{modality type, polarity, propositional arguments, source, background, modified element, degree indicator, outscoping quantifier} and \textit{lemma}. The types into which modality was categorized are: fine grained types: \textit{epistemic, circumstantial, ability, deontic, bouletic, teleological, and bouletic/teleologica}l; and the coarse grained types: \textit{epistemic or circumstantial, ability or circumstantial, and priority}.

On surveying other studies carried out using annotations, apparently it seems that more and more attributes were annotated to make text understanding precise. Certain works moved in the direction of subjectivity analysis; some in certainty analysis; whereas others focused on time, events and temporal analysis; all of them being implicitly useful in the study of type and/or sense understanding of modality.

Different types of subjectivity is implied in discourse by different types, epistemic and deontic modals. The relationship between subjectivity and modality is elaborately discussed in \citet{sanders1997perspective}.

Rubin et al. \citeyearpar{rubin2004certainty,rubin2006certainty} presented a certainty categorization model based on four hypothesized dimensions and tested the model on a sample of news articles. \citet{rubin2010epistemic} identifies that certainty can be seen as a variety of epistemic modality expressed in form of markers like \textit{probably, perhaps, undoubtedly}, etc.

\citet{nissim2013cross} proposed modality annotation model that they say to be two layered. Factuality and speaker’s attitude being two components marked, they also plan to make the model more coherent by annotating strength of modality.

\citet{matsuyoshi2010annotating} clearly draws attention to the point that recent developments in language processing has improved precision but is insufficient for applications such as information extraction, question answering and recognizing textual entailment. Such applications require more information such as modality, polarity, and other associated information collectively referred as \textit{extended modality}. Matsuyoshi proposes an annotation scheme that represents extended modality and consists of seven components: \textit{source, time, conditional, primary modality type, actuality, evaluation,} and \textit{focus}. Utilizing the work, they also constructed an annotated corpus in Japanese. 

Emphasizing upon event modality, Saur\'{\i} et al. \citeyearpar{sauri2006annotating} says that modality is an important component of discourse together with other levels of information such as argument structure and temporal information. That made an apparent need for a more sophisticated approach that is sensitive to such additional information. They worked on annotation scheme that annotates event modality and also identifying its scope using TimeML. Saur\'{\i} \citeyearpar{sauri2006slinket} has also worked on SlinkET attempting the construction of modal parser for events. Pustejovsky et al. \citeyearpar{pustejovsky2003timebank} built the TimeBank corpus which is annotated with information like modals, events, times, relation between events and temporal expressions. \citet{sauri2009factbank} built the FactBank on the basis of TimeBank, where events are assigned with different degrees of factuality according to their source-introducing predicates (SIPS) and source. Different degrees of factuality are determined by different degrees of certainty and polarity axes. Degrees of certainty include \textit{certain, probable, possible} and \textit{unknown}. Whereas polarity axis contained \textit{positive, negative} and \textit{underspecified}. 

\section{Gap and Futurology}
\label{Gap}

Linguistic modality is one of the components of discourse that is associated with context, sense and meaning, mental and real spaces, and force dynamics.

Also there is no proper well defined classification of different types and senses of modality in linguistic literature. This lack of taxonomy has contrived enormous confusions regarding types and senses. Considering the fact that the section is complex and modality in discourse has many aspects associated with it, and its wider scope, even language scientists can contribute towards it.

First point noted is that different kinds of expressions can be used to convey modal meanings. And we saw that the tagging approach is limited to tag modals’ explicit use in text. Moreover, taggers don\textquotesingle t put any further light on the type and sense of the modality. Though attempts have made on type and sense classification using annotation methods, but due to flexibility of meanings, it is difficult to standardize. Flexibility of meaning means the modal verb has different meaning according to context. Modality in language has contextual meanings and implied perceptions. And mechanisms fail to identify perspective, aspect and contextuality.

Another drawback is dependency in both methods; tagging as well as annotating. The process of tagging is dependent on the tagset and the annotation scheme is limited by its own training corpus. Although this lexicon dependency of the available approaches are useful for preliminary passes of processing but not an effective way for understanding natural language which includes cognition and perception.

At this point, inspiring from human information processing mechanisms would be seen appropriate. In this context understanding of human cognitive process can definitely enlighten the path of development to make our systems artificially intelligent. Computational models of cognition, creative insight, skill acquisition and the design of instructional software, as well as other topics in higher cognition needs to be reconsidered. It is noteworthy to see the human’s perceptive systems, visual or speech/audio, etc., both are essentially layered and hierarchical in structure. Thus, it is natural to believe that the state-of-the-art can be advanced in processing these types of natural language if structurally efficient and effective learning model can be developed. The layered structure of human learning shown in Kaplan \& Sadock's Comprehensive Textbook of Psychiatry \cite[fig. 24-1]{sadock2000kaplan} is instrumental in visualization of complexity and multi-stage structure involved. Moreover, the interconnection and synergy between the layers is equally vital.

Real systems are dynamic in nature and reform continuous shift that produce perceptual difference. If the shift is in upward direction (in hierarchical multi-layered model) that results in high dimensionality in nature of expression. Thus the expression becomes nonspecific and losses subjectivity. Nonspecific discourse are too complex to interpret and to reach to any conclusion is very difficult. From viewing the software dealing with language processing looks in direction and trends accordance to the subject Artificial Intelligence (AI) and Robotics that attempts to mimic the dynamic and behavioral output from human. In context of natural language processing with special reference to modality processing noticeable development observed in static format of expressions and expression of contextuality also attempted within one document as in MMAX2 from German NLP group.

Neural network hidden layer processing and continuous modification were successfully executed in mechanical output in field of machine learning. The directions were explored with annotated titles like deep learning and layered approach handling are most popular among research works in language processing. 

As per the objectives of a system, some key features that should be considered such as; relevant well-structured knowledge base with improved feature space should be formed upon each processing. And this knowledge base must be dynamically updated (active learning that means continuous updatation of the knowledge base by each layer of the model) so that it can effectively be useful in applications of the system.

Machine learning has been a dominant tool in NLP for many years. However, the use of traditional machine learning in NLP has been mostly limited to numerical optimization of weights for human designed representations and features from the text data. The goal of representation learning is to automatically develop features or representations from the raw text material appropriate for a wide range of NLP tasks.

Deep learning is gaining popularity very recently as it provides levels of abstraction. The multi-layered architecture formed due to the levels ensure natural progression from low level to high level structure as seen in natural complexity. Deep learning works on the principle of formation of learning representations. The essence of deep learning is to automate the process of discovering effective features or representations for any machine learning task, including automatically transferring knowledge from one task to another concurrently. In regard to NLP, deep learning develops and makes use an important concept called “embedding”, which refers to the representation of symbolic information in natural language text at word-level, phrase-level, and even sentence-level in terms of continuous-valued vectors.

Another concept of multi-task learning has also shown improvements in learning approaches. Multi-task learning is a machine learning approach that learns to solve several related problems at the same time, using a shared representation. It can be regarded as one of the two major classes of transfer learning or learning with knowledge transfer, which focuses on generalizations across distributions, domains, or tasks. The other major class of transfer learning is adaptive learning, where knowledge transfer is carried out in a sequential manner, typically from a source task to a target task.

\subsection{Recent works with deep architectures in NLP}
Variety of deep architectures like neural networks, deep belief networks and others has shown significant performance in various applications of language processing including other fields.

\cite{collobert2011natural} provide a comprehensive review on ways of applying unified neural network architectures and related deep learning algorithms to solve NLP problems from “scratch”, meaning that no traditional NLP methods are used to extract features. The recent work by \cite{mikolov2013efficient} derives word embeddings by simplifying the Neural Network Language Model (NNLM). It is found that the NNLM can be successfully trained in two steps. Yet another deep learning approach to machine translation appeared in  \cite{mikolov2013exploiting}. 

One most interesting NLP task recently tackled by deep learning methods is that of knowledge base (ontology) completion, which is instrumental in question-answering and many other NLP applications. An early work in this space came from \cite{bordes2011learning}, where a process is introduced to automatically learn structured distributed embeddings of knowledge bases. The proposed representations in the continuous-valued vector space are compact and can be efficiently learned from large-scale data of entities and relations. A specialized neural network architecture is used. In the follow-up work that focuses on multi-relational data \citep{bordes2014semantic}, the semantic matching energy model is proposed to learn vector representations for both entities and relations. 

Other recent works \citet{socher2013reasoning} and \citet{bowman2013can}, adopts an approach, based on the use of neural tensor networks, to attack the problem of reasoning over a large joint knowledge graph for relation classification. The knowledge graph is represented as triples of a relation between two entities, and the authors aim to develop a neural network model suitable for inference over such relationships. The model they presented is a neural tensor network, with one layer only but it would be encouraged to work further on multi-layered network models. The network is used to represent entities in fixed-dimensional vectors, which are created separately by averaging pre-trained word embedding vectors. It then learns the tensor with the newly added relationship element that describes the interactions among all the latent components in each of the relationships. Experimentally, Socher et al., shows that this tensor model can effectively classify unseen relationships in WordNet and FreeBase. Thus, models built on tensors can contribute upto certain extent for reasoning over relationships between entities enhancing knowledge bases for improved performance. Works utilizing Recursive Neural Networks (RNN) for syntactic parsing and word representations has been performed in \cite{luong2013better,socher2010learning}. Deep neural networks have been popular and are well performing as they are intrinsically multi-layered in structure. 

Deep learning is a hot area of research and there is still much potential for significant advances. It can be said that the paradigm of deep learning architectures can improve the results of our models upto quite a certain extent; but there is still a limit to it considering the whole problem statement. This is because the deep neural networks are yet a kind of black box model in terms of functionality. By revising models and designs enhancement in performance of deep learning algorithms can surely be made as per specific application domain but there would be a bound to the possible improvements and the available approaches wouldn’t provide enough means to the desired level of Artificial Intelligence. Approaches that are multi-layered and preferably white box models would be essentially important for organization and control of intermediate layers’ functionalities.

As mentioned above, not only for application in NLP but for any application that deals with dynamics of real world conditions, development of complex system using combination of several simple modular and multilayered hierarchal architectures with the key features will be helpful. And hopefully such models can attain better linguistic understanding accuracy that would be contributory to the application expertise.





\bibliographystyle{spbasic.bst}
\bibliography{mybib.bib}







\end{document}